\documentclass[runningheads]{llncs}
\usepackage{graphicx}
\usepackage{subfigure}
\usepackage{amsmath,amssymb} % define this before the line numbering.
\usepackage[width=122mm,left=12mm,paperwidth=146mm,height=193mm,top=12mm,paperheight=217mm]{geometry}
\usepackage{color}
\usepackage{array}

\usepackage{algorithm}
\usepackage{algorithmic}
\usepackage{setspace}
\newcommand{\ve}[1]{\mathbf #1}
\usepackage{epsfig}
\usepackage{array}
\usepackage{float}
\usepackage{caption}
\usepackage{enumitem}
\begin{document}
\pagestyle{headings}
\mainmatter

%\def\ACCV16SubNumber{159}  % Insert your submission number here

%===========================================================
\title{Weighted Residuals for Very Deep Networks} % Replace with your title
\titlerunning{Weighted Residuals for Very Deep Networks}
\authorrunning{Shen \emph{et al.}}

\author{Falong Shen, Gang Zeng}
\institute{Peking University}

\maketitle
\begin{abstract}
Deep  residual networks have recently shown appealing performance on many challenging computer vision tasks. However, the original residual structure still has some defects making it difficult to converge on very deep networks.
In this paper, we introduce a weighted residual network to address the incompatibility between \texttt{ReLU} and element-wise addition and the deep network initialization problem. The weighted residual network is able to learn to combine residuals from different layers effectively and efficiently. The proposed models enjoy a consistent improvement over accuracy and convergence with increasing depths from 100+ layers to 1000+ layers. Besides, the weighted residual networks have little more computation and GPU memory burden than the original residual networks.
The networks are optimized by projected stochastic gradient descent.
Experiments on CIFAR-10 have shown that our algorithm has a \emph{faster convergence speed} than the original residual networks and reaches a \emph{high accuracy} at 95.3\% with a 1192-layer model.
\end{abstract} 
\section{Introduction}
 The state-of-the-art model for image classification is built on inception and residual structure \cite{russakovsky2015imagenet,szegedy2016inception,he2015deep}. Lots of works devoted on residual networks are emerging recently \cite{he2016identity,targ2016resnet,huang2016deep,wide2016deep}. Very deep convolutional networks \cite{krizhevsky2012imagenet,simonyan2014very}, especially with residual units, have shown compelling accuracy and nice convergence behaviors on many challenging computer vision tasks \cite{he2015deep,dai2015instance,ren2015faster}. Since \emph{vanishing gradients} problem is well handled by batch normalization \cite{ioffe2015batch} and highway signal propagation \cite{srivastava2015training}, networks with 100+ layers are being developed and trained, even 1000+ layers structure still yields meaningful results when combined with adequate dropout as shown in \cite{huang2016deep} . He \emph{et al.} \cite{he2016identity} also introduced the pre-activation structure to allow the highway signal to be directly propagated through the very deep networks. However they seemed to harness the features with a larger dimension (4$\times$) and adapted multiple $1 \times 1$ convolutional layers to substitute $3 \times 3$ convolutional layers for convergence with 1000+ layers.

% ----------------------------- point 1 -----------------------------
 A typical convolutional unit is composed of one convolutional layer, one batch normalization layer and one \texttt{ReLU} layer, all of which are performed sequently \cite{ioffe2015batch}. For a residual unit, a central question is how to combine the residual signal and the highway signal, where element-wise addition was proposed in \cite{he2015deep}. A natural idea is to perform addition after \texttt{ReLU} activation. However, this leads to a non-negative output from residual branch, which limits the representative ability of the residual unit meaning that it can only enhance the highway signal.
 He \emph{et al.} \cite{he2015deep}  firstly proposed to perform addition between batch normalization and \texttt{ReLU}. In \cite{he2016identity}, they further proposed to inverse the order of the three layers, performing batch normalization and \texttt{ReLU} before convolutional layers.
 The question is due to that \texttt{ReLU} activation can only generate positive value which is incompatible with element-wise addition in the residual unit.
% ----------------------------- point 2 -----------------------------

 As it is non-convex optimization to solve deep networks, an appropriate initialization is important for both faster convergence and a good local minima. The ``xavier'' \cite{glorot2010understanding} and ``msra'' \cite{he2015delving} are popular used for deep networks initialization. However, for networks with depths beyond 100 layers, neither ``xavier''  nor ``msra'' works well. The paper of \cite{he2015deep} proposed to ``warm up'' the network with small learning rate and then restore the learning rate to normal value. However, this hand-craft strategy is not that useful for very deep networks, where even a very low learning rate (0.00001) still is not enough to promise convergence and restoring the learning rate has a chance to get rid of the initial convergence \cite{szegedy2016inception}.

Generally speaking, there are two defects embedded in the training of the original residual networks
\begin{itemize}
  \item Incompatibility of \texttt{ReLU} and element-wise addition.
  \item difficutly for networks to converge with depths beyond 1000-layer using ``msra'' initializer.
\end{itemize}
% ----------------------------- point 3 -----------------------------
The third point resides that a better mode to combine the residuals from different layers are necessary to train very deep networks. For very deep networks, not all layers are that important as 1000-layer networks often perform not much better than 100-layer networks. In fact, lots of layers serve as redundant information and very deep networks tend to over-fit on some tasks.

% ----------------------------- contribution -----------------------------------------------------
 In this paper, we introduce the weighted residual networks, which learn to combine residuals from different layers effectively and efficiently. All the residual weights are initialized at zeros and optimized with a very small learning rate (0.001), which allows all the residual signals to gradually  add to the highway signal. With a group of gradually growing-up residual weights, the 1192-layer residual networks converge even much faster than the 100-layer networks. Finally, the distribution of the learned residual weights is in a symmetry mode ranging in $[-0.5,0.5]$, which implies the incompatibility of \texttt{ReLU} and element-wise addition can be appropriately handled.  The networks are optimized by projected stochastic gradient descent with exactly the same training epochs to original residual networks.

 We conduct experiments on CIFAR-10 \cite{krizhevsky2009learning} to verify the practicability of the weighted residual networks. Training with the weighted residual networks can \emph{converge much faster} and reach a \emph{higher performance} with \emph{negligible more computation and GPU memory cost} than the original residual networks. The weighted residual networks with depths beyond 1000 layers still converge faster than shallower networks and enjoy a consistent improvement over accuracy with increasing depths from 100+ layers to 1000+ layers without resorting to any hand-craft strategy such as ``warm up'' \cite{he2015deep}. After applying dropout on the residuals, our weighted residual networks reach a very high accuracy (95.3\%) on CIFAR-10 using a 1192-layer model with the same training epochs to the original residual networks (about 164 epochs, 64k iterations).

The contributions of our work presented in this paper have four folds:

\begin{itemize}
  \item We propose the weighted residual networks, which learn to combine the residuals from each residual unit. The weighted residual networks converge much faster in the training stage and reach a higher accuracy than the original residual networks at little more computation and GPU memory cost.
  \item The incompatibility of \texttt{ReLU} and element-wise addition can be addressed appropriately by weighted residuals and we clear all the obstacles on the information highway to allow the highway signal to enjoy a unhindered propagation.
  \item The  residuals are gradually added to the highway signal to make the training process more reliable, even networks with depths beyond 1000 layers can converge very fast without the ``warm up'' strategy.
  \item We modify the down-sampling step to make the spatial size and feature dimension consistent  between highway signal and branched residual signal, without resorting to zero-padding or extra converting matrix.
\end{itemize}
 The weighted residual networks are simple and easy to implement while having surprising practical effectiveness, which makes it particular useful for complicated residual networks in research community and real applications.

\section{Related works}
The residual networks have attracted lots of researchers and many works on it have appeared \cite{he2016identity,targ2016resnet,huang2016deep,wide2016deep,shah2016deep}. In the following paragraphs we will review some related works.

The residual networks simplify the highway networks \cite{srivastava2015training} using identity skip connection, which allows information to flow directly and bypass complex layers. The residual networks consist of many residual units. There are two information flows in a residual unit. The highway signal goes through the identity skip connection  and the branched residual signal is realized by \texttt{Conv-BN-ReLU-Conv-BN}. The two flows are combined at the end of a residual unit by element-wise addition and then it goes through a \texttt{ReLU} layer for activation. This simple structure is quite powerful and achieved a surprising performance on the imageNet challenge \cite{russakovsky2015imagenet}  with 150-layer networks \cite{he2015deep}.
%------------------------------------------------------ paper original -------------------------------------------

In the original residual networks, the two flows are added up before \texttt{ReLU} activation for a numerical reason that \texttt{ReLU} can only produce \emph{non-negative output}, which means the branched residual signal can only enhance the highway signal. However, intuitively it is not a natural solution as the branched residual signal needs to be ``activated''.

%------------------------------------------------------ paper pre-activation -------------------------------------------
He \emph{et al.} \cite{he2016identity} proposed to handle the incompatibility between \texttt{ReLU} and element-wise addition by re-arranging these layers to \texttt{BN-ReLU-Conv-BN-ReLU-Conv} and named it ``pre-activation'' structure. When applying the ``pre-activation'' structure, special attention should be taken on the first and the last residual unit of the networks.

%------------------------------------------------------ paper dropout -------------------------------------------
To train ``residual'' networks, it is natural to fit on the ``residual'' only, which means when the branched residual signal  is not presented, the highway signal should still make meaningful results. Under this condition, the branched residual signal can focus on fitting the ``residual'' in a residual unit. Huang \emph{et al.} \cite{huang2016deep} proposed a dropout residual network, which randomly drops the branched residual signal in each residual unit. Therefore, when the  branched residual signal is presented in a residual unit, it can focus on fitting the ``residual''. As this model can be treated as an ensemble of models with different depths, they named it ``stochastic depth networks''.

%------------------------------------------------------- paper feature dimension --------------------------------------

In the convolutional networks, the depth and width are both important for a high performance in image classification \cite{wide2016deep,he2015deep}. The \texttt{conv1-conv3-conv1} bottleneck structure which used a feature dimension $4\times$ larger than \texttt{conv3-conv3} reached a higher performance \cite{he2016identity}. Zagoruyko \emph{et al.} \cite{wide2016deep} used \texttt{conv3-conv3} with feature dimension $10\times$ larger and reached the highest performance on CIFAR-10 (4.10\%). However, a larger feature dimension costs much more GPU memory and leads a shallower structure. There is a balance between depth and width.

In this paper, we mainly focus on models with depth beyond 100+ layers. We mean to explore how to train a very deep model effectively instead of tuning a more accurate model. 
\section{Weighted Residual Networks}

\begin{figure}
  \centering
  \includegraphics[width=1\textwidth]{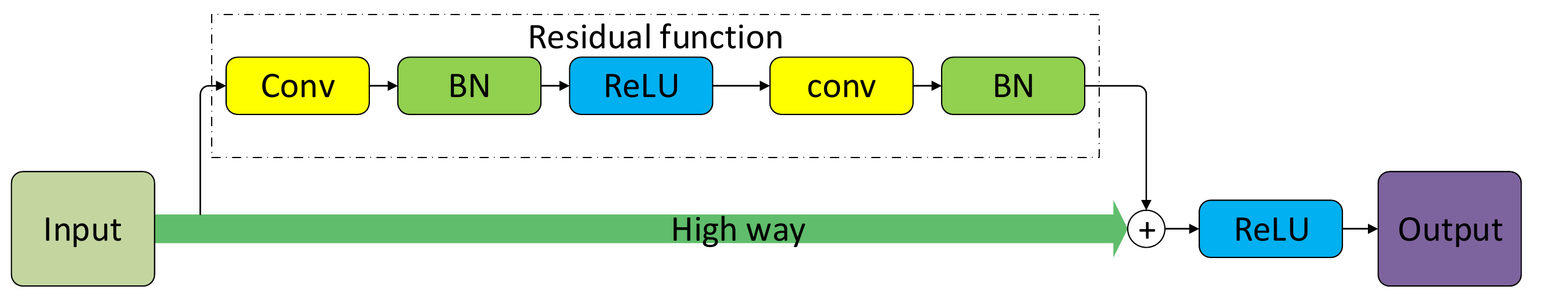}
  \caption{Diagrammatic sketch of a residual unit. The residual function comprises of two $3 \times 3$ convolution layers. Each Convolutional layer (\texttt{Conv}) is followed by a batch normalization layer (\texttt{BN}) and a \texttt{ReLU} layer (\texttt{ReLU}). The weights of convolutional layers are initialized by ``msra''. The highway signal and the residual signal are combined by element-wise addition.}
  \label{figure:head_2}
\end{figure}

Firstly we will give a brief introduction to the residual networks. The residual networks build the information highway by allowing earlier feature representation to flow unimpededly and directly to the following layers \emph{without} any modification.  A residual unit performs the following computation:
\begin{equation}\label{equation:start}
   \ve x_{i+1} = \texttt{ReLU}(\ve x_i + \Delta L_i(\ve x_i,\theta_i))
\end{equation}
Here $\ve x_i$ is the input highway signal to the $i$-th residual unit. $\theta_i$ is the  filter parameters for the residual unit and it is initialized by ``msra'', $\Delta L_i$ is the residual function, which is realized by a stack of two $3 \times 3$ convolutional layers. Typically, one convolutional layer should be followed by one batch normalization layer to keep the signal with non-zero variance  and one \texttt{ReLU} layer for non-linearity activation. The highway should be clean and unhindered. As it is shown in \cite{he2016identity}, obstacles on the highway, such as constant scaling and dropout, will make the optimization difficult. A typical residual unit is depicted in Figure \ref{figure:head_2}.

The original residual networks stated above have two defects
\subsubsection*{Incompatibility of \texttt{ReLU} and element-wise addition.}

 The highway signal and the residual signal which is produced by the residual function are combined by the element-wise addition. However, the element-wise addition is operated between the \texttt{BN} layer and the \texttt{ReLU} layer after the second \texttt{Conv} layer. This is mainly due to the \texttt{ReLU} activation function, which produces \emph{non-negative output}. The output of \texttt{ReLU} operation is not compatible with element-wise addition as it can only enhance the highway signal, which limits the representability of the residual function, which is meant to take values in $(-\infty,+\infty)$. One can of course resort to designing other activation function which can take values in a larger range or a symmetry mode around zero.

\subsubsection*{Initialization of very deep networks.}

Very deep networks with depths beyond 1000 layers, even equipped with residual structure, batch normalization and \texttt{ReLU}, still do not converge in the training stage as shown in Figure \ref{figure:4figure}. The paper of \cite{he2015deep} proposed to ``warm up'' the network training with a little learning rate for several epochs and then restore it to the  normal learning rate in order to facilitate the initial convergence. However, for deeper networks, even very little learning rate may not work well \cite{szegedy2016inception}.

In very deep networks, the residuals from each block are added together and make the training hard to converge. One may want to zero all the residuals to start the training. However, \emph{the weights of the convolutional layers} in residual functions should be initialized by ``msra''  which has little probability to produce all-zero weights.

%-------------------------------------------------------------------------------------------------------------------
\subsection{Weighted Residuals}
To address the incompatibility of \texttt{ReLU} and element-wise addition and to get a better initialization for very deep networks, we introduce the weighted residual networks. Formally in a weighted residual networks unit, the computation of the signal is
\begin{equation}\label{start}
    \ve x_{i+1} = \ve x_i + \lambda_i\Delta L_i(\ve x_i,\theta_i), \lambda_i \in (-1,1),
\end{equation}
where $\theta_i$ is the filter parameters and it is initialized by ``msra'' , $\lambda_i$ is the weight scalar for the residual and it is initialized by zero with a very small learning rate. The \texttt{ReLU} activation is removed from the highway and  $\Delta L_i$ is realized by two \texttt{Conv-BN-RelU}s.

\begin{figure}
  \centering
  \includegraphics[width=1\textwidth]{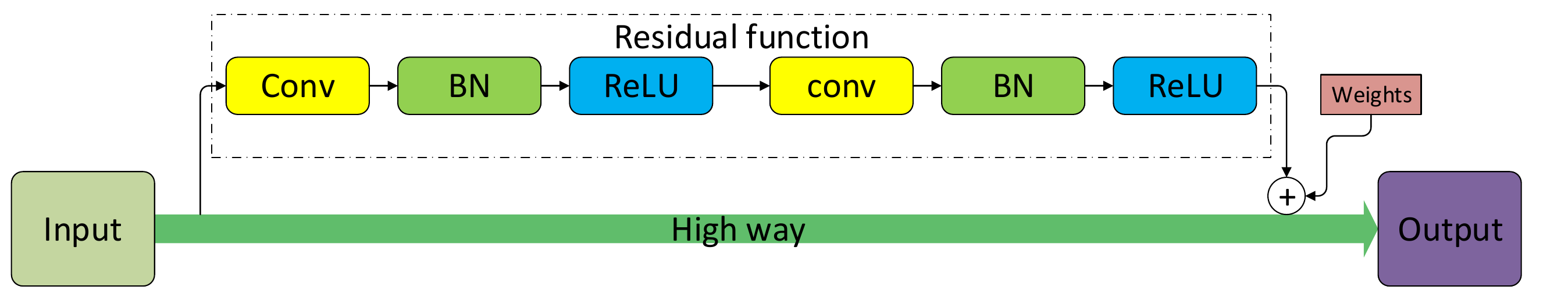}
  \caption{Diagrammatic sketch of a weighted residual unit. We move the \texttt{ReLU} from highway to the branchway, which allows the highway signal to flow unobstructedly through the very deep networks. The residual signal is weighted by a scalar which is initialized by zero in the training stage. In our experiments, the overall convergence is promised when all the residuals are gradually added to the highway signal. The weight takes values in $(-1,1)$ to overcome the limitation of the \texttt{ReLU} activation function.}
  \label{figure:head_1}
\end{figure}

For any deep blocks, the feature representation $\ve x_{i+k}$ in the $(i+k)$-th layer can be expressed as a summation of the input layer representation $\ve x_i$ and a series of weighted residual functions,
\begin{equation}
    \ve x_{i+k} = \ve x_i + \sum_{j=1}^{k}\lambda_{i+j}\Delta L_{i+j}(\ve x_{i+j},\theta_{i+j}), \lambda_{i+j} \in (-1,1).
\end{equation}
In the back-propagation stage, the gradient of any layer does not vanish when filter parameter $\theta_{i+j}$ is \emph{arbitrarily small}. Note that the pre-activation structure proposed in \cite{he2016identity} also has a similar property by converting the order of \texttt{Conv-BN-RelU} to \texttt{BN-RelU-Conv}.

\begin{figure}
  \centering
  \includegraphics[width=1\textwidth]{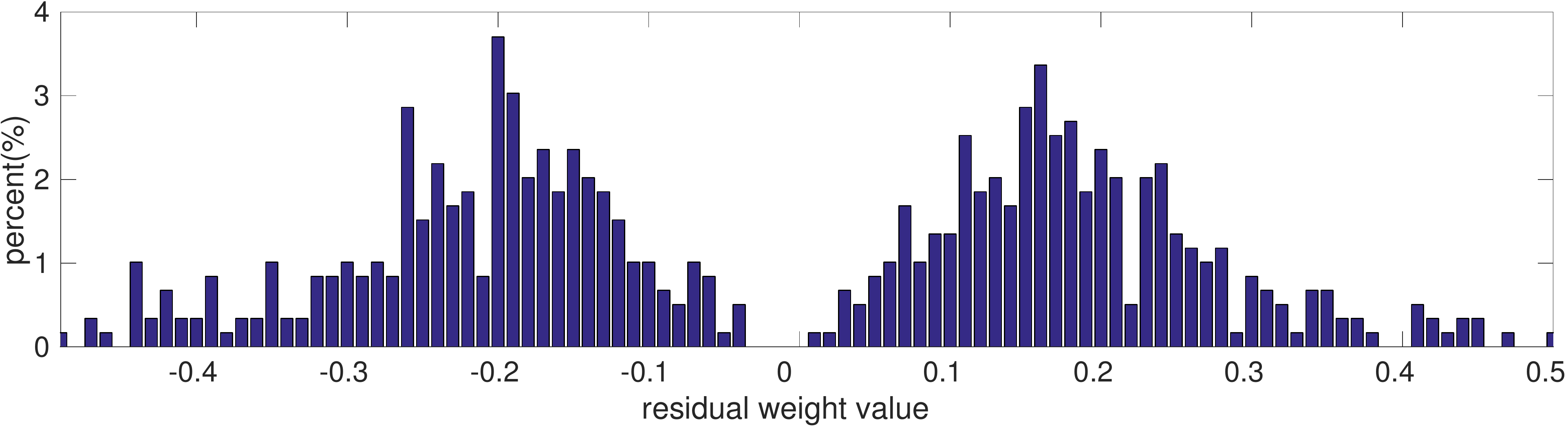}
  \caption{Distribution of the learned residual weights in a 1192-layer model.}
  \label{figure:weigts_dist}
\end{figure}

In Figure \ref{figure:weigts_dist} we visualize the distribution of the learned residual weights in a 1192-layer model. The residual weight values range around (-0.5,0.5) in a symmetry mode, which means the branched residual signal has equal probability to enhance/weaken the highway signal, which means the incompatibility between \texttt{ReLU}  and element-wise addition is appropriately addressed by the learned residual weights.

\vspace{-0.8cm}

\subsection{Modification to the structure.}
At the beginning of a new block in the original residual networks, the highway signal is down-sampled by a stride-2 convolution layer while the branched residual signal also need to be halved by a stride-2 convolution layer. When performing the element-wise addition, zeros-padding or convert matrix is necessary to make a matched feature dimension between the two signals.

\begin{figure}
  \centering
  \includegraphics[width=1\textwidth]{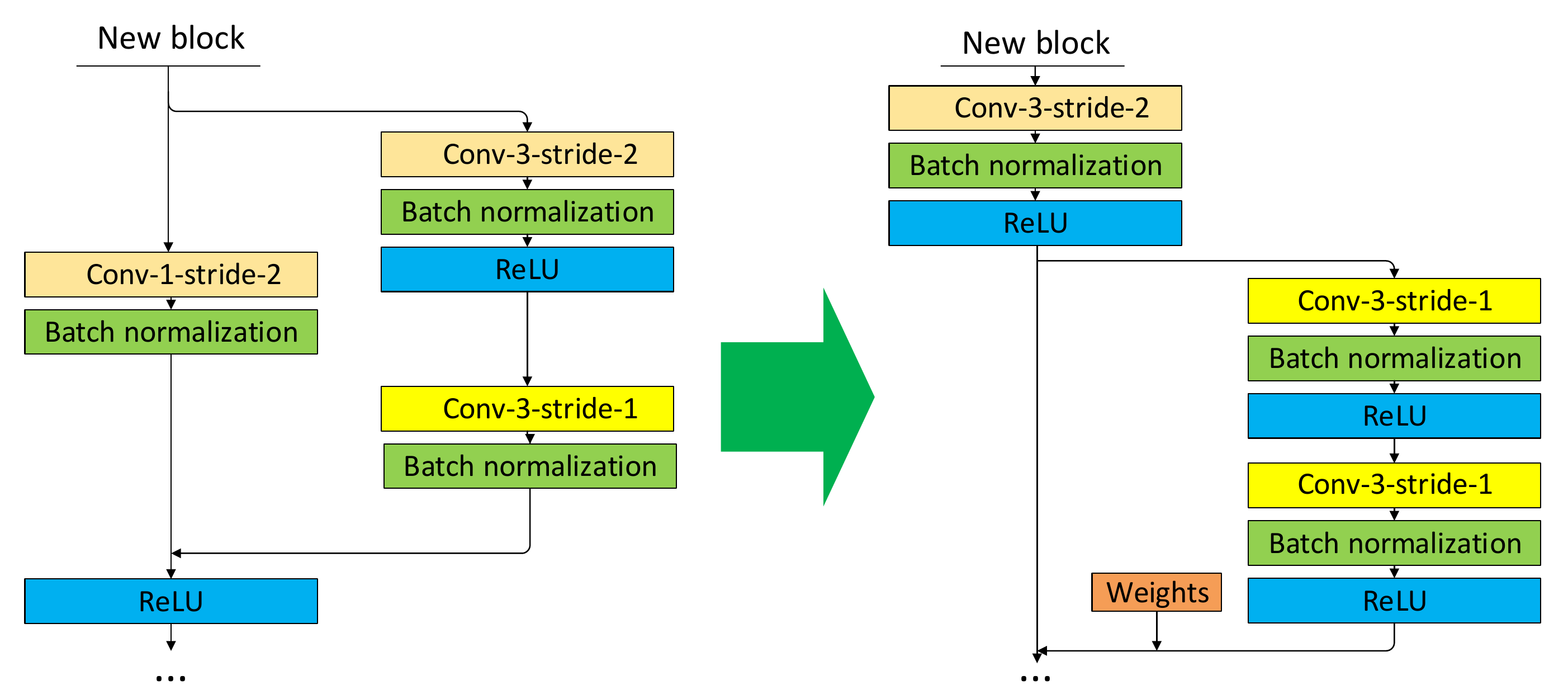}
  \caption{At the beginning of each new block, the feature map is halved by a \texttt{conv-3-stride-2} layer.}
  \label{figure:stride}
\end{figure}

In our networks as it is shown in Figure \ref{figure:stride}, we directly halve the feature size at the beginning and the following layers are performed as stated in the previous sections.

\subsection{Optimization.}
 Given training images  and its corresponding ground truth labels $\{I_i,y_i\}$, the loss function is the summation of the negative likelihood and the regularized term
\begin{equation}\label{loss}
    \begin{split}
  -\frac{1}{|\mathcal{I}|}\sum_{i \in \mathcal{I}}\log p_{y_i}(I_i,\theta,\lambda)+\frac{1}{2}||\theta||_2^2+\frac{1}{2}||\lambda||_2^2, \\
  s.t. ~~each~elemenet~of~  \lambda \in (-1,1).
  \end{split}
\end{equation}
where $\theta$ is the network parameters which is initialized by ``msra'', $\lambda$ is the weight vector for the residuals and is initialized by all-zeros. We apply projected \texttt{SGD} to this typical constraint optimization problem. In the $(t+1)$-th iteration, the updated  $\lambda^{t+1}_i$ is projected to the convex set $S$
\begin{equation}\label{lambda}
  \lambda^{t+1}_i=P_S(\lambda^t_i+\Delta\lambda^t_i)
\end{equation}
where the convex set $S=(-1,1)$ and $\Delta\lambda^t_i$ is the gradient of the loss function in Equation \ref{loss} with regard to $\lambda^t_i$, which is effectively computed  by back-propagation \cite{hecht1989theory} in deep networks.

\subsection{Implementation details.}

\subsubsection*{Dataset.}
CIFAR-10 \cite{krizhevsky2009learning} is a dataset of color images all coming with the same size of $32 \times 32$, which consists of 50k training images and 10k testing image in 10 classes. We train our deep model on the \emph{train} set and evaluate the finally trained models on the \emph{test} set.  We follow the same residual architecture as proposed in \cite{he2015deep}.

Our code is built on the open source deep learning framework Caffe \cite{jia2014caffe}. We use a weight decay of 0.0001 and momentum of 0.9 with batch size of 128. The initial learning rate is 0.1 without ``warm up'' for any model. The initial learning rate for residual weights is set to 0.001. The filter parameters are initialized by ``msra'' \cite{he2015delving}. The residual weights are set to all-zeros.  We are not meant to push the state-of-the-art performance on CIFAR-10 so we follow the same training strategy as \cite{he2015deep}. All the models are trained for 64k iterations and the learning rate is divided by 10 at 32k and 48k iterations.  We also adapt the simple data augmentation as it is shown in \cite{lee2014deeply}: 4 pixels are padded around the training images with zero-values and a translated or mirrored $32\times32$ crop is fed into the networks. We do not have \emph{val} set and the model at the end of training is used to perform on the \emph{test} set. In the test stage, the original $32\times32$ images are evaluated.

The network contains three blocks  and the feature map is halved twice. There are totally $6n+4$ layers as it is shown in Table \ref{table:layers}. We compare $n=\{1, 3,9,18,48,100,198\}$, which leads to 10, 22, 58, 112, 292, 604 and 1192-layer networks.

\begin{table}[H]
\centering
\caption{Our weighted residual network comprises of three blocks similar to \cite{he2015deep}. The first block is started with a $3 \times 3$ convolutional layer with stride 1. The latter two blocks are started with a $3 \times 3$ convolutional layer with stride 2. There are $6n+4$ layers in the whole networks (with the final \texttt{FC} layer).}
\vspace{3mm}
\begin{tabular}{p{2cm}<{\centering}|p{2cm}<{\centering}|p{2cm}<{\centering}|p{2cm}<{\centering}}
  \hline
  feature size & 32 $\times$ 32 & 16 $\times$ 16 & 8 $\times$ 8  \\
  \hline
  filter number & 16         & 32            & 64 \\
  \hline
  layer number & 2n+1        & 2n+1          & 2n+1 \\
  \hline
\end{tabular}
\label{table:layers}
\end{table} 
\section{Experiments}
In this section we present and analyze the experiment results on CIFAR-10 to demonstrate the effectiveness of the weighted residual networks.

\begin{figure}[H]
    \subfigure[training curve]
    {
        \begin{minipage}[b]{0.475\textwidth}
            \includegraphics[width=1\textwidth]{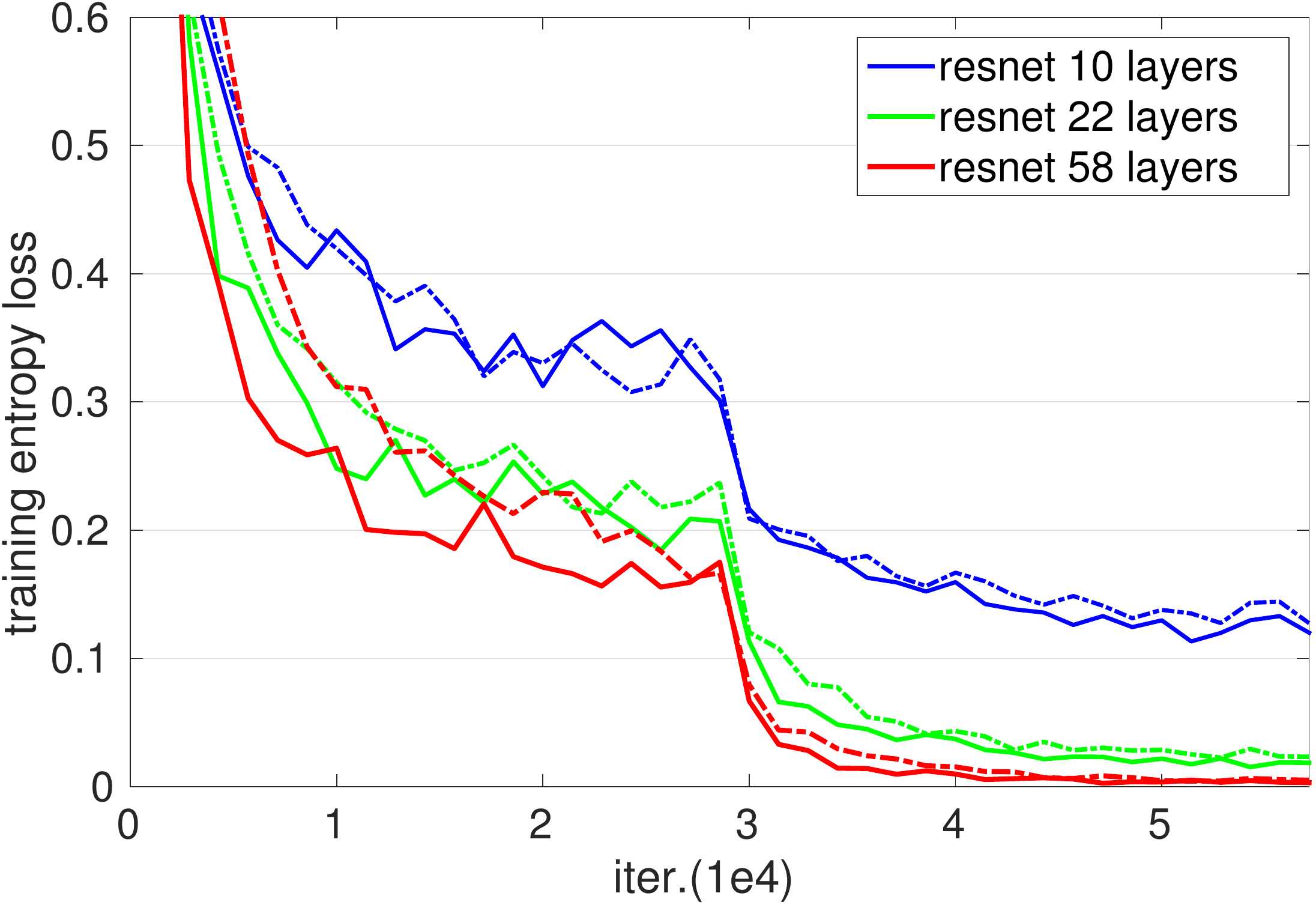}
        \end{minipage}
    }
    \subfigure[test curve]
    {
        \begin{minipage}[b]{0.475\textwidth}
            \includegraphics[width=1\textwidth]{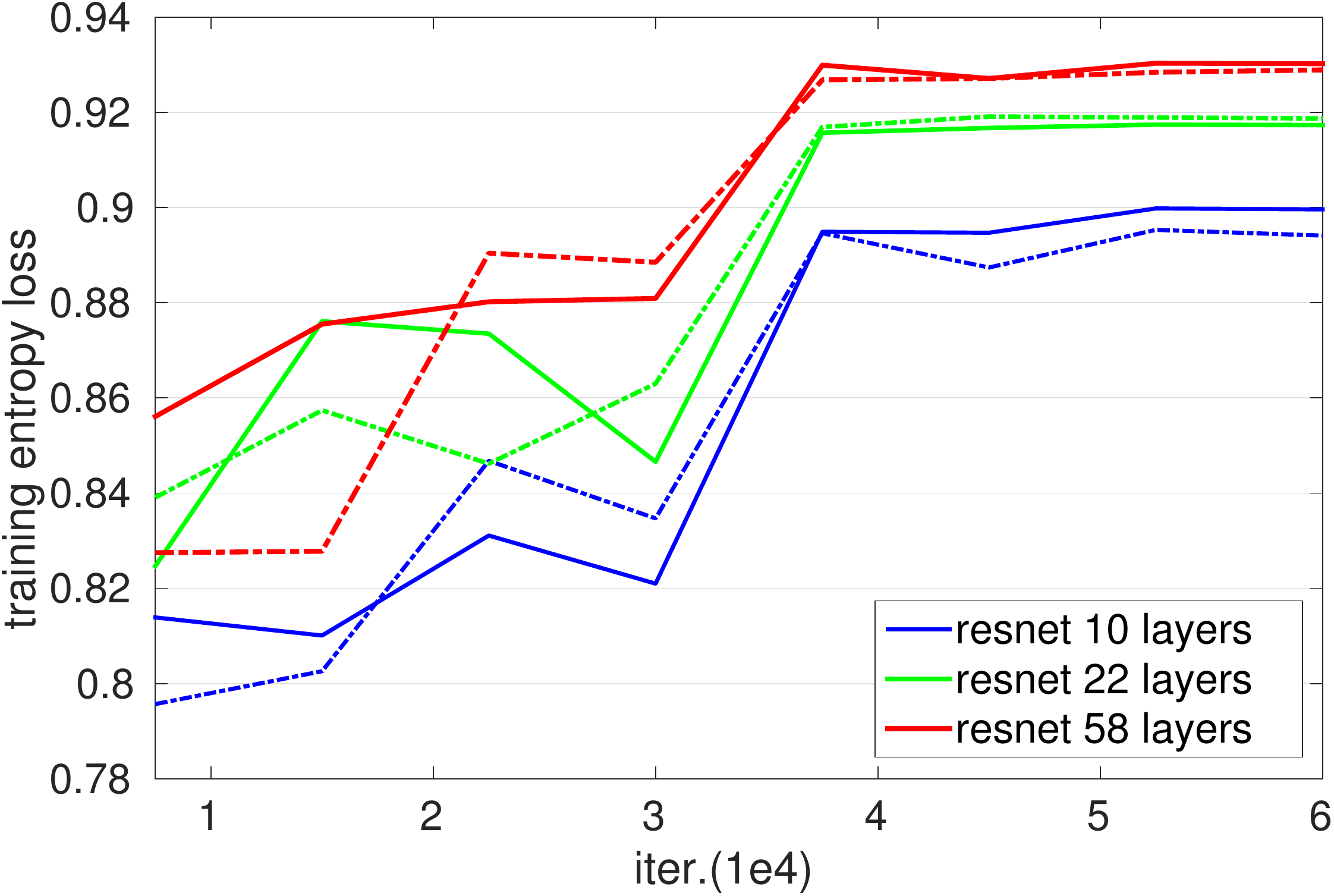}
        \end{minipage}
    }
    \subfigure[training curve]
    {
        \begin{minipage}[b]{0.475\textwidth}
            \includegraphics[width=1\textwidth]{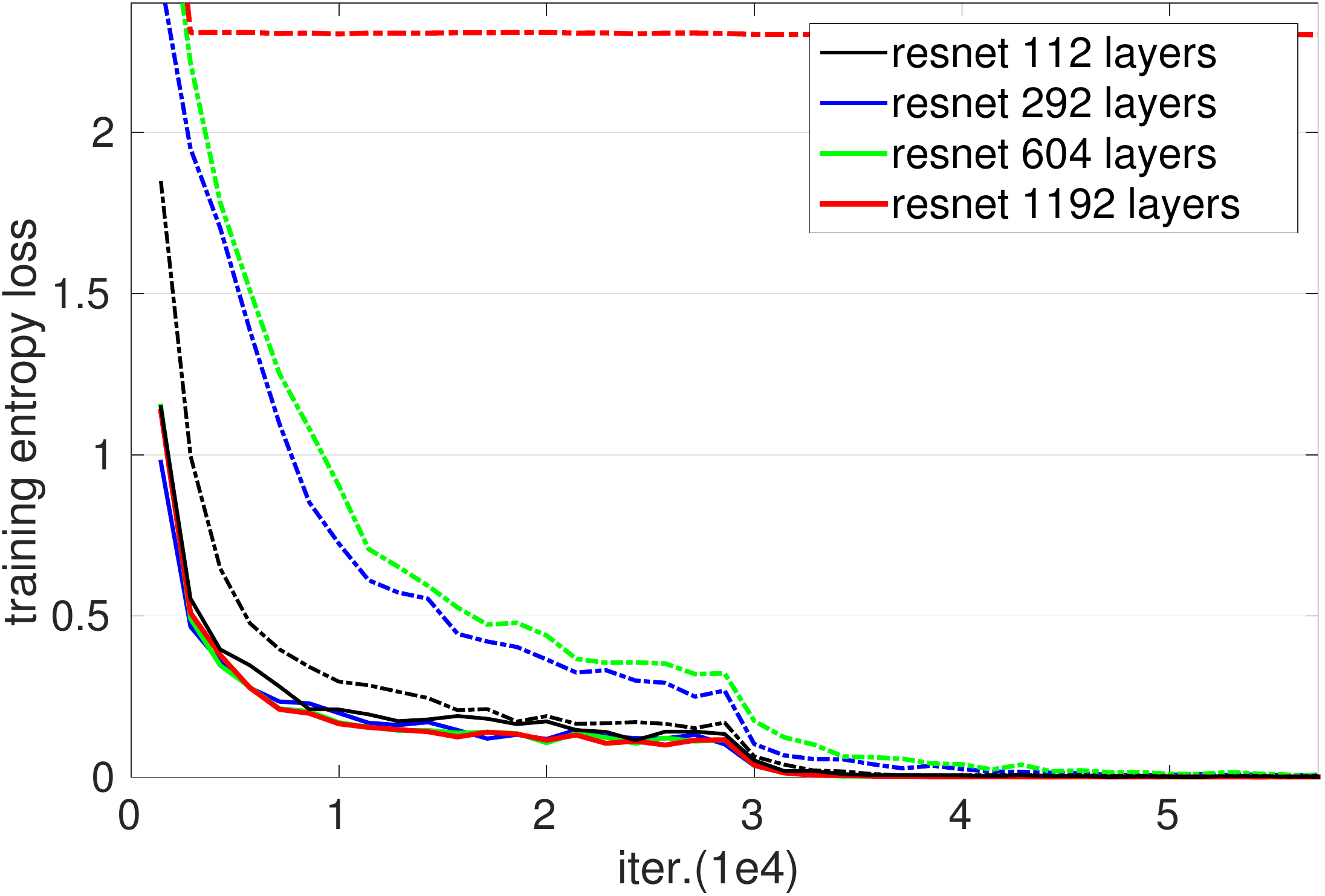}
        \end{minipage}
    }
    \subfigure[zoom in at training curve]
    {
        \begin{minipage}[b]{0.475\textwidth}
            \includegraphics[width=1\textwidth]{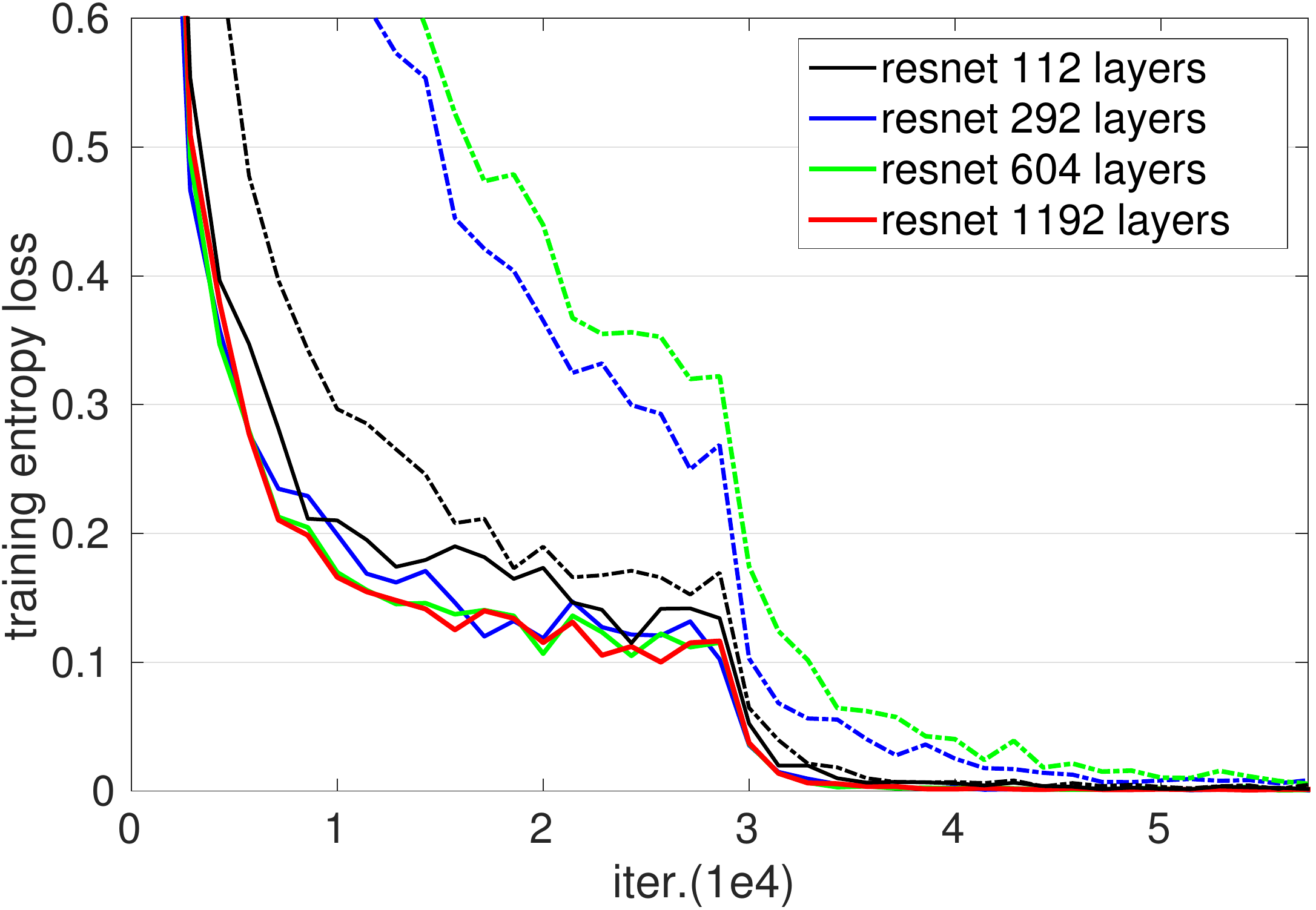}
        \end{minipage}
    }
    \vspace{-0.3cm}
    \caption{Comparisons of the original residual networks and the weighted residual networks on the CIFAR-10. The \texttt{bold lines} denote the weighted residual networks and the \texttt{dashed lines} denote the original residual networks. The \textbf{top left} figure is the training entropy loss and the \textbf{top right} figure is the corresponding test accuracy on shallow networks. The \textbf{bottom left} figure records the training entropy loss for very deep networks and the \textbf{bottom right} figure is the zoomed version for more details.}
    \label{figure:4figure}
\end{figure}
\vspace{-1.6cm}
\begin{figure}[H]
  \centering
  \includegraphics[width=0.8\textwidth]{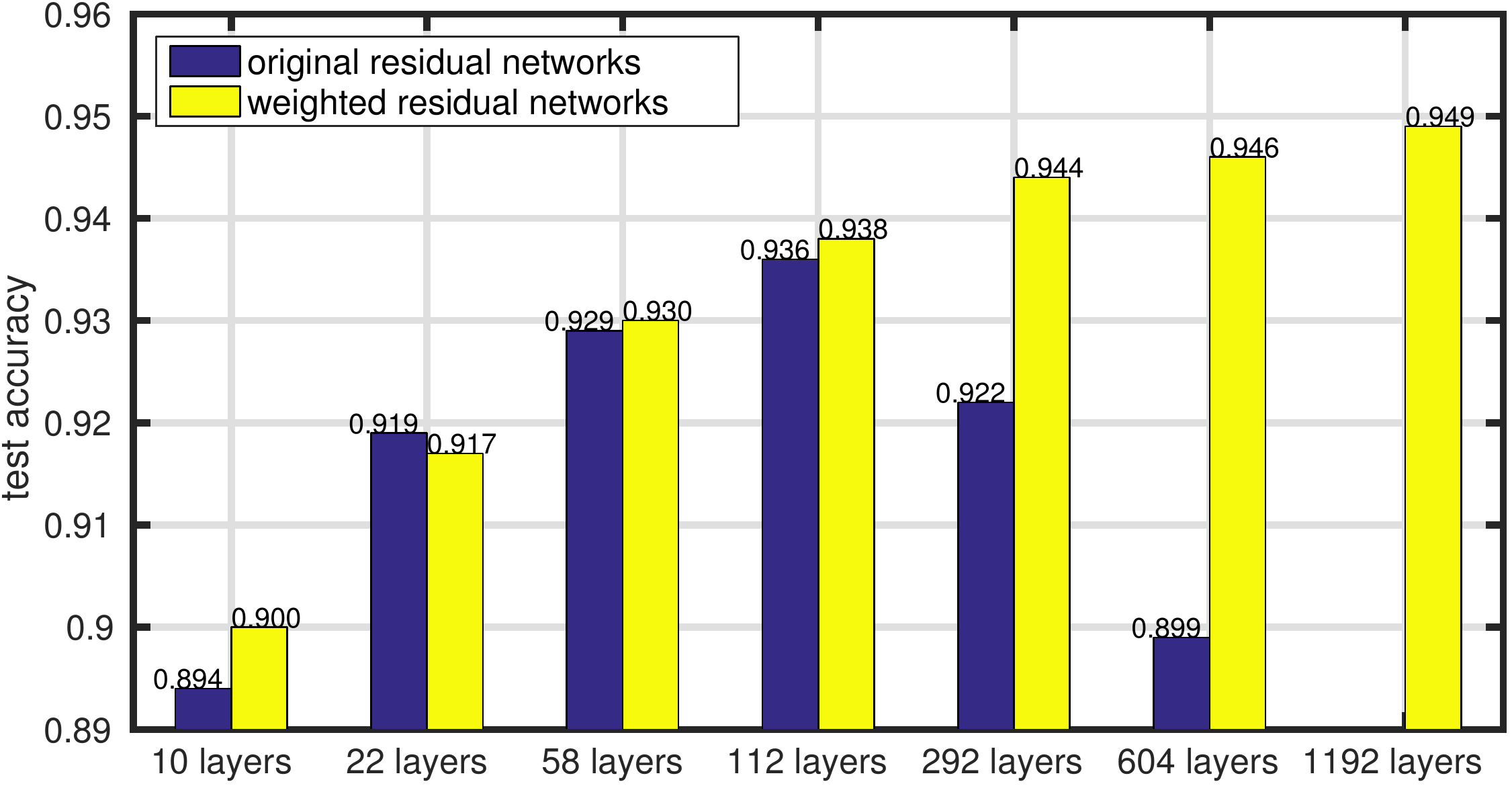}
  \caption{Test accuracy on CIFAR-10. The original 1192-layer residual network fails to reach a meaningful result in the training stage and we do not report it.}
  \label{figure:test_all}
\end{figure}

\subsection{Results}
\textbf{Convergence.} Firstly we experiment on shallow networks (layer number $<$ 100). As it is shown in Figure \ref{figure:4figure}(a) and Figure \ref{figure:4figure}(b), both of the weighted residual networks and the original residual networks have very similar performance of convergence and final accuracy on shallow networks.

Then we conduct experiments on very deep networks (layer number $>$ 100). In Figure \ref{figure:4figure}(c), the weighted residual network shows much better performance on convergence in the training stage. In fact, networks with depths beyond 1000 layers still converge faster than the 112-layer networks in Figure \ref{figure:4figure}(d). As contrary, the original residual network does not converge well and the 1192-layer network even does not converge at all as we did not apply the ``warm up'' strategy. However, even equipped with ``warm up'', the original 1192-layer residual network ends with over-fitting and reaches a worse performance than the 112-layer network  as it is reported in \cite{he2015deep}.

\noindent\textbf{Accuracy.} The overall test accuracy of deep networks on CIFAR-10 is reported in Figure \ref{figure:test_all}. The \texttt{blue histograms} denote the performance of the original networks. The accuracy decreases after the layer number is larger than 100. However, for the weighted residual networks, which are denoted as \texttt{yellow histograms}, the performance enjoys a consistent improvement with the increasing depths from 10+ layers to 1000+ layers.

The weighted residual networks can always \emph{converge faster} and reach a \emph{higher performance} when there are \emph{more layers} throughout our experiments.

\subsection{Comparison with the State-of-the-art}
In this subsection we compare the  weighted residual networks (\textbf{WResNet}) with other recently proposed models. Mainly there are two kinds of models, first of which focus on enlarging the feature dimension and we call them \texttt{wide models}, the second of which focus on depths and we call them \texttt{deep models}. Note that 1001-layer Pre-activation \cite{he2016identity} is both deep (1000+ layers) and wide ($4\times$ feature dimension)  model. The results are presented in Table \ref{table:othermodels}. All these models, except for Highway \cite{srivastava2015training}, share similar structures with ResNet \cite{he2015deep}, including three feature blocks.

\begin{table}
\caption{Test accuracy(\%) on CIFAR-10.}
\begin{tabular}{p{2.0cm}<{\centering}|p{2.2cm}<{\centering}|p{1.5cm}<{\centering}|p{2.0cm}<{\centering}|p{1.5cm}<{\centering}|p{1.5cm}<{\centering}}
  \hline
   Type           & method          & depth & feature dim. & epochs & acc.(\%) \\
  \hline
                    &Highway              & 32        & -            & -                 &  91.2            \\
                  &Pre-activation       & 164         & 64-128-256           & 200              &   94.5            \\
   wide models    &WideDim              & 28          & 160-320-640          & 200               &   95.8             \\
                  &RiR                 & 18         & 96-192-384            & 82                  &  95.0             \\
  \hline
                 &ResNet               & 1202         & 16-32-64               & 164             &    92.1          \\
                 &Pre-activation       & 1001         & 64-128-256           & 200             &   95.4            \\
   deep models    &Dropout              & 1202         & 16-32-64            & 300                  &  95.1             \\
                 &\textbf{WResNet}              & 1192         & 16-32-64           & 164              &  94.9            \\
                 &\textbf{WResNet-d}              & 1192         & 16-32-64           & 164              &  95.3            \\
  \hline
\end{tabular}
\label{table:othermodels}
\end{table}

Pre-activation \cite{he2016identity} adapted a \texttt{conv1-conv3-conv1} bottle-neck structure and enlarged the feature dimension by $4\times$. Apparently a $4\times$ wider model enjoys a higher performance but costs more GPU memory. As the GPU memory (12GB for one GTX TITAN X) resource is limited,  it is important to tune the model width and depth economically for a very accurate model. WideDim \cite{wide2016deep} and RiR \cite{targ2016resnet} are two other methods to enlarge the feature dimension for higher accuracy. A clear tendency is that a wider feature is better for higher performance. WideDim adapted a $10\times$ feature dimension and reached a very high performance (95.8\%) on CIFAR-10. Dropout \cite{huang2016deep} realized stochastic depth networks by applying the dropout operation on the residual signal at exactly the same GPU memory cost. The only defect resides that it needs much more epochs (about $2\times$) to converge at a good performance.

The weighted residual networks make very deep networks training converge faster and reach a good performance while bringing little more computation and GPU memory burden. As time and GPU resource is limited, we have not tuned the model width (feature dim.) or more training epochs and we are meant to explore the effectiveness of the weighted residuals in \emph{training very deep models}.  Yet with shorter feature dim., the weighted residual networks still perform much better than the original residual networks and reach a quite meaningful accuracy as shown in Table \ref{table:othermodels}.

We further apply dropout on the residuals with \emph{dropout\_ratio} = \{0.2,0.4,0.6\} for three blocks as proposed by \cite{huang2016deep}. The performance of this model is named as \textbf{WResNet-d}. With only about half training epochs of \cite{huang2016deep}, the weighted residual networks with dropout reach a relative very high performance (95.3\%).

\subsection{Analysis}
We provide more insights into the weighted residual networks by presenting more details information of results in this subsection.
\begin{figure}
    \includegraphics[width=1\textwidth]{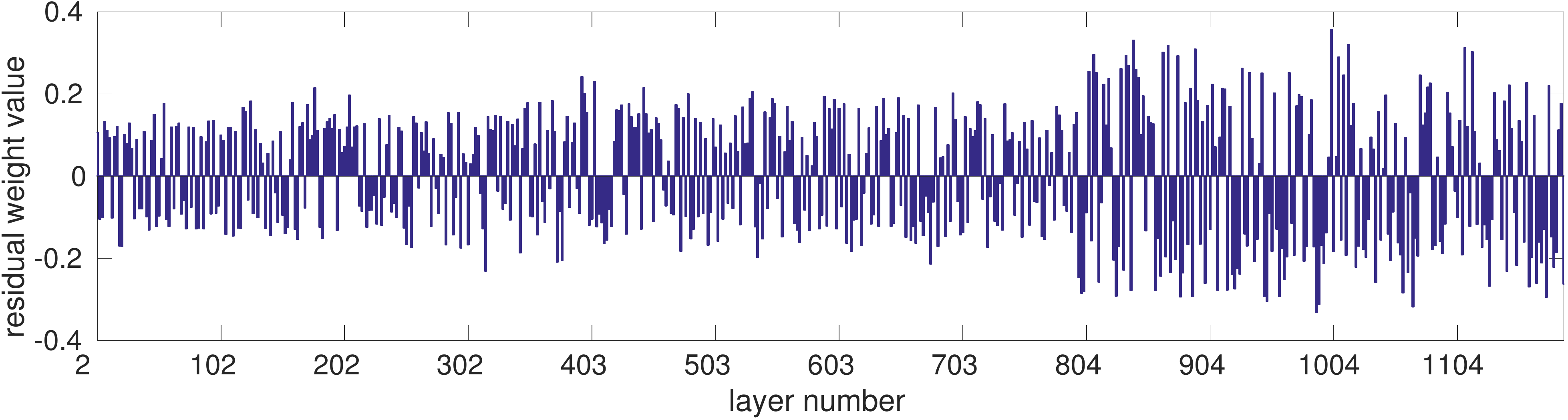}
    \caption{The residual weight values from a trained 1192-layer model on CIFAR-10.}
    \label{figure:weight}
\end{figure}

\begin{figure}
    \subfigure[8k iterations]
    {
        \begin{minipage}[b]{1\textwidth}
            \includegraphics[width=1\textwidth]{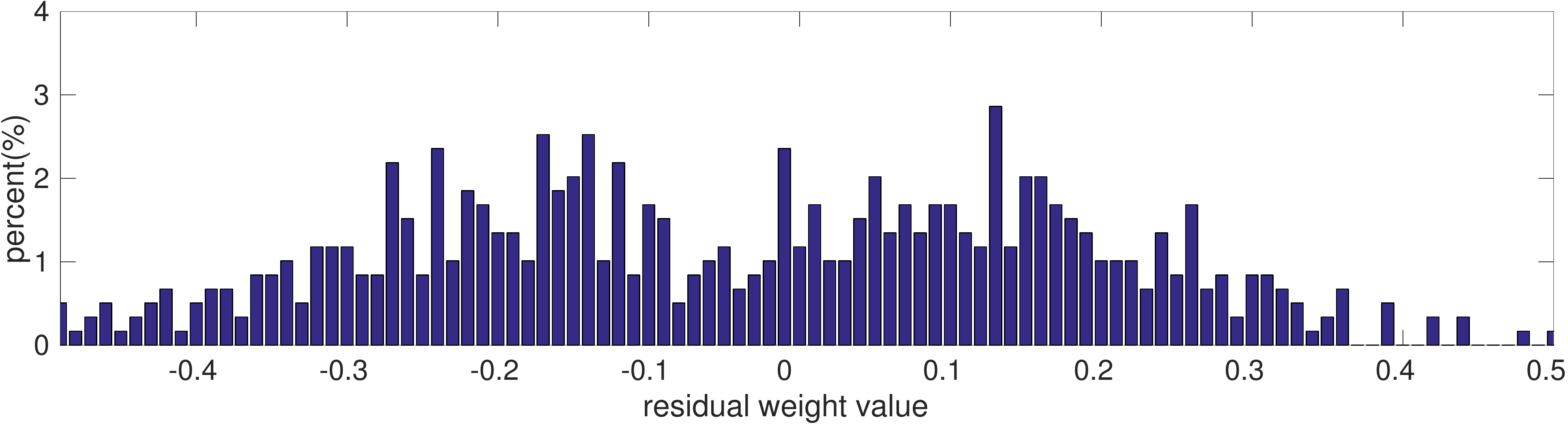}
        \end{minipage}
    }
    \subfigure[16k iterations]
    {
        \begin{minipage}[b]{1\textwidth}
            \includegraphics[width=1\textwidth]{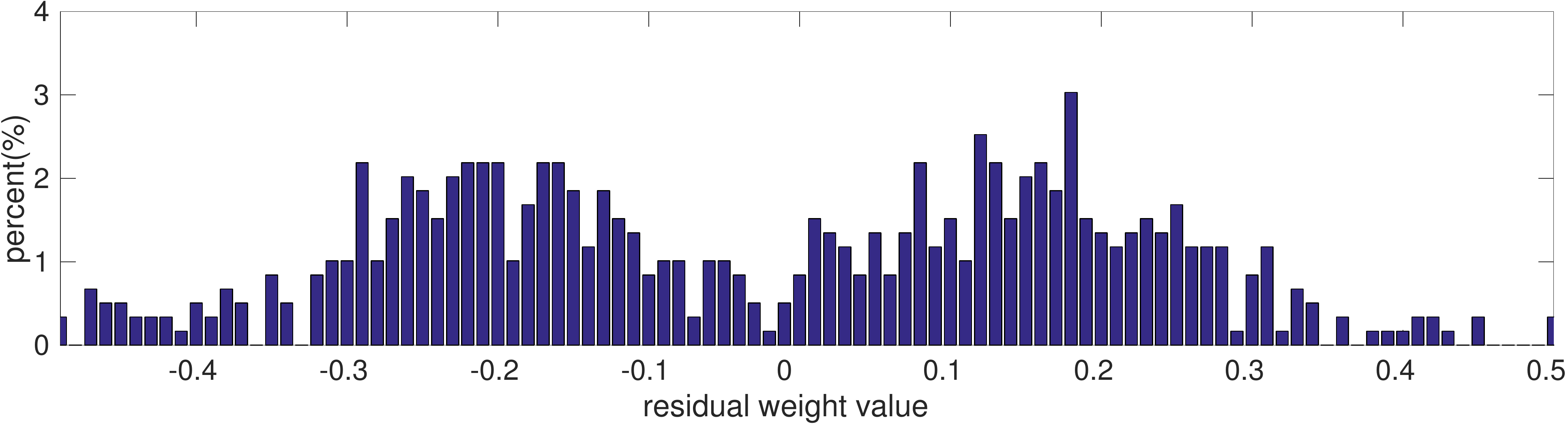}
        \end{minipage}
    }
    \subfigure[32k iterations]
    {
        \begin{minipage}[b]{1\textwidth}
            \includegraphics[width=1\textwidth]{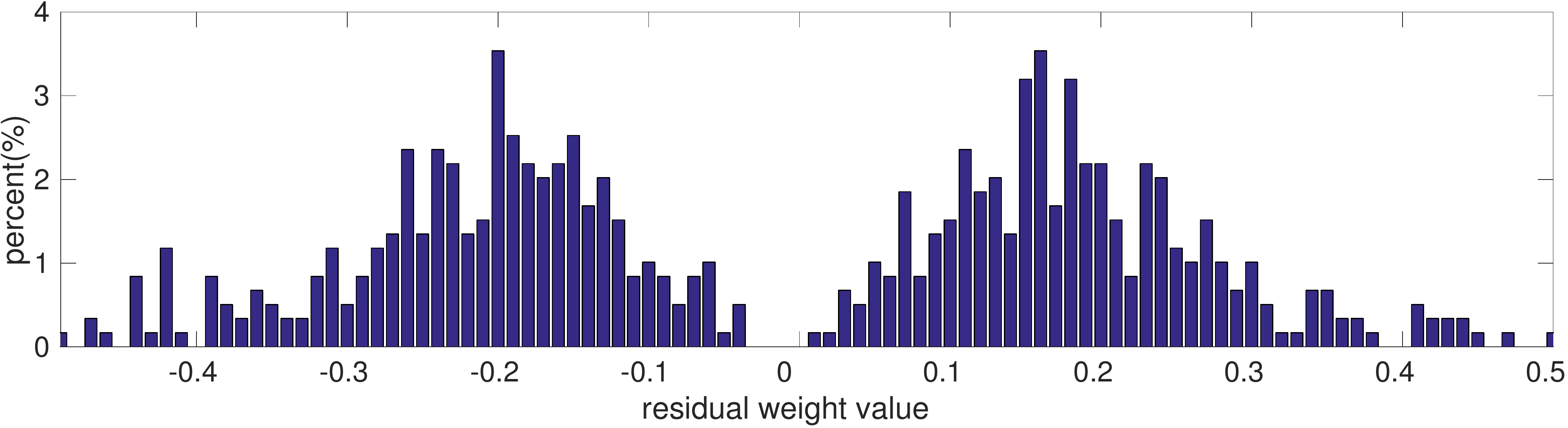}
        \end{minipage}
    }
    \subfigure[64k iterations]
    {
        \begin{minipage}[b]{1\textwidth}
            \includegraphics[width=1\textwidth]{figures/hist_64000.pdf}
        \end{minipage}
    }
    \caption{Evolution of the distribution of the residual weight values in the training stage of a 1192-layer weighted residual network on CIFAR-10.}
    \label{figure:evolution}
\end{figure}

The initial learning rate for the residual weights is set to 0.001 for all models and the residual weights are initialized with zeros. Figure \ref{figure:weight} shows the learned residual weight values in each element-wise addition layer in a 1192-layer model. It comprises two parts divided by a visible sharp boundary around the 800-layer and the latter residuals have larger weights. It may imply the residuals from the later layers are more important than earlier layers on the final decisions. We will explore this phenomenon in the future work.

We have also plotted the evolution history of the distribution of the residual weight  values as show in Figure \ref{figure:evolution}. At the 8k iteration, the distribution is relative uniform. As more and more training iterations, the distribution begins to concentrate around two peaks. In the 64k iteration, most of the residual weight values are around $0.2$ and $-0.2$ in a symmetry mode indicating that the branched residual signals have equal probability to enhance/weaken the highway signals, which verifies our hypothesi. Therefore the learned residual weights can solve the incompatibility between \texttt{ReLU} activation and element-wise addition appropriately.

\section{Conclusion}
The original residual networks have two defects, 1) Incompatibility between \texttt{ReLU} and element-wise addition. 2) Difficulty for networks to converge with depths beyond 1000-layer using ``msra'' initializer. In this paper we introduce the weighted residual networks to make very deep residual networks \emph{converge faster} and reach a \emph{higher performance} with \emph{little more computation and GPU memory burden} than the original residual networks. All the residuals are added to the highway signal gradually by the learned slowly growing-up weights to promise convergence. Experiments on CIFAR-10 have demonstrated the effectiveness of the weighted residual networks for very deep models. It enjoys a consistent improvements over accuracy and convergence with the increasing depths from 100+ layers to 1000+ layers.
The weighted residual networks are simple and easy to implement while having surprising practical effectiveness, which makes it particular useful for complicated residual networks in research community and real applications.

\bibliographystyle{splncs}
\bibliography{egbib}

%this would normally be the end of your paper, but you may also have an appendix
%within the given limit of number of pages
\end{document}